\documentclass{article}
\usepackage{spconf,amsmath,graphicx}
\usepackage{numprint}
\usepackage{xspace}						
\usepackage{extdash}    
\usepackage{xcolor}
\usepackage{subcaption} 

\makeatletter \def\argmax{\mathop{\operator@font argmax}}
\makeatletter \def\argmin{\mathop{\operator@font argmin}}

\makeatletter
\DeclareRobustCommand\onedot{\futurelet\@let@token\@onedot}
\def\@onedot{\ifx\@let@token.\else.\null\fi\xspace}

\def\eg{\emph{e.g}\onedot}

\def\etal{\emph{et al}\onedot}

\usepackage[marginclue,footnote]{fixme}
\fxsetup{theme=color,mode=multiuser,status=draft}
\FXRegisterAuthor{am}{aam}{\color{blue}AM}
\FXRegisterAuthor{bl}{abl}{\color{orange}BL}
\FXRegisterAuthor{cr}{acr}{\color{red}CR}

\title{TOWARD RELIABLE MODELS FOR AUTHENTICATING MULTIMEDIA CONTENT: DETECTING RESAMPLING ARTIFACTS WITH BAYESIAN NEURAL NETWORKS}
\name{Anatol Maier, Benedikt Lorch, Christian Riess\thanks{This work was supported by Deutsche Forschungsgemeinschaft (DFG, German Research Foundation) as part of the Research and Training Group 2475 ``Cybercrime and Forensic Computing'' (393541319/GRK2475/1-2019).
This material is based on research sponsored by the Air Force Research
Laboratory and the Defense Advanced Research Projects Agency under agreement
number FA8750-16-2-0204. The U.S. Government is authorized to reproduce and
distribute reprints for Governmental purposes notwithstanding any copyright
notation thereon. The views and conclusions contained herein are those of the
authors and should not be interpreted as necessarily representing the official
policies or endorsements, either expressed or implied, of the Air Force
Research Laboratory and the Defense Advanced Research Projects Agency or the
U.S. Government.
}}
\address{IT Security Infrastructures Lab, Friedrich-Alexander University Erlangen-N{\"u}rnberg, Germany}
\begin{document}
\ninept
\noindent
\textcopyright~2020 IEEE. Personal use of this material is permitted. Permission from IEEE must be obtained for all other uses, in any current or future media, including reprinting/republishing this material for advertising or promotional purposes, creating new collective works, for resale or redistribution to servers or lists, or reuse of any copyrighted component of this work in other works.
\newpage
\maketitle
\begin{abstract}
In multimedia forensics, learning-based methods provide state-of-the-art performance in determining origin and authenticity of images and videos. However, most existing methods are challenged by out-of-distribution data, i.e., with characteristics that are not covered in the training set. This makes it difficult to know when to trust a model, particularly for practitioners with limited technical background.

In this work, we make a first step toward redesigning forensic algorithms with a strong focus on reliability. To this end, we propose to use Bayesian neural networks (BNN), which combine the power of deep neural networks with the rigorous probabilistic formulation of a Bayesian framework. Instead of providing a point estimate like standard neural networks, BNNs provide distributions that express both the estimate and also an uncertainty range.

We demonstrate the usefulness of this framework on a classical forensic task: resampling detection. The BNN yields state-of-the-art detection performance, plus excellent capabilities for detecting out-of-distribution samples. This is demonstrated for three pathologic issues in resampling detection, namely unseen resampling factors, unseen JPEG compression, and unseen resampling algorithms.  We hope that this proposal spurs further research toward reliability in multimedia forensics.
\end{abstract}
\section{Introduction}
\label{sec:intro}

The goal of image forensics is to validate origin and authenticity of digital
images.
%
Most state-of-the-art forensic methods are based on deep learning. Notable
example applications are to blindly validate noise
statistics~\cite{cozzolino2019noiseprint}, to link EXIF tags to noise
statistics~\cite{huh2018exifConsistency}, to detect artifacts of commonly used
operators~\cite{bayar2018processingOperations,boroumand2018processingHistory}, to detect
computer-generated imagery~\cite{yu19ganFingerprints,
roessler2019faceForensics}, or to detect JPEG
inconsistencies~\cite{barni2017aligned}.

The success of learning-based methods can broadly be attributed to their
excellent capability in deriving most subtle cues from training
examples. However, one notable disadvantage of learning-based
methods is their sensitivity to out-of-distribution data: if a test image 
differs too much from the training dataset, learning-based methods oftentimes
exhibit difficulties in detecting these cues~\cite{cozzolino2018forensicTransfer}.

Such out-of-distribution scenarios are, unfortunately, quite common in the
practical forensic work: oftentimes, little is known about the exact provenance
of an image, particularly when it was found on the internet. For example, there
may be only limited knowledge about the type and amount of in-camera processing
that an image underwent, and limited knowledge about distribution-related
postprocessing such as the implementation of the JPEG library of a web
platform. Current forensic methods address this issue with extensive data augmentation~\cite{cozzolino2019noiseprint, huh2018exifConsistency, yu19ganFingerprints, roessler2019faceForensics}. The aim of such data augmentation is to
anticipate a rich variety of commonly seen processing steps and compression variants to
harden the network against variations of the input data.

However, although current methods achieve impressive results on a wide variety
of inputs, augmentation can only extend the horizon of \textit{seen data}. It
does not provide knowledge about potential failure cases from \textit{unseen
data}. This can be a severe practical limitation: a forensic analyst has to
understand on which data a method can operate or not in order to assess its output.

In this work, we propose a different direction to mitigate this issue:
\textbf{The first contribution} is to propose Bayesian neural networks (BNNs)
for image forensics to intrinsically model uncertainty about the data. BNNs
combine the strengths of deep neural networks with a rigorous probabilistic
framework. Thus, BNNs are also amenable to data augmentation to increase the
range of seen data, but they can also detect whenever operating on unseen data.
This enables an analyst to know about network uncertainty without requiring
expertise in the technical specifics.

\textbf{The second contribution} is to demonstrate the usefulness of the
proposed approach on a classical forensic task, namely the detection of
resampling. The proposed BNN achieves state-of-the-art detection performance.
Additionally, it shows impressive capabilities to detect out-of-distribution
inputs on three notorious issues in resampling detection, namely previously
unseen rescaling ranges, JPEG postcompression, and rescaling algorithms.

This paper is organized as follows. In Sec.~\ref{sec:related_work}, we
describe related work on uncertainty modeling in neural networks and on
resampling detection. Section~\ref{sec:methods} introduces the basic concepts
of the Bayesian framework for convolutional neural networks (CNNs) and
variational inference. Section~\ref{sec:experiments} presents experimental
results on various resampling scenarios. Section~\ref{sec:conclusion}
concludes with a brief summary and outlook.

\section{Related Work}
\label{sec:related_work}

Detecting out-of-distribution examples has recently gained popularity in the machine learning community.
Hendrycks and Gimpel proposed the softmax activations of a neural network to
anticipate incorrect classifications and detect out-of-distribution
samples~\cite{hendrycks17softmaxStatisticsBaseline}. However, the usefulness of
softmax statistics is limited. Guo~\etal show that these softmax probabilities
do not accurately represent the ``true correctness
likelihood''~\cite{guo2017calibration}. 
Liang \etal~\cite{liang2018odin} propose to calibrate softmax activations to
the model confidence via temperature scaling and input preprocessing similar to
the fast gradient sign method introduced by~\cite{goodfellow2014explaining}.
%
%
%
This approach results in a local distillation of the input space, but it does
not solve the general problems that neural networks make confident errors and
do not provide information of their predictive uncertainty.
To address these issues, DeVries and
Taylor~\cite{deVries2018confidencefromHints} propose to learn confidence
estimates by training a network with two output branches producing prediction
and uncertainty estimate.

Similar goals to model uncertainty can be achieved with a Bayesian framework,
which has a more solid theoretical foundation. Gal and
Ghahramani~\cite{gal2016dropout} show how Bayesian inference can be
approximated with standard CNNs using dropout at test time. Inspired by their
approach, Lakshminarayanan \etal~\cite{lakshminarayanan2017deepEnsembles} use
an ensemble of networks to obtain uncertainty estimates. Both works, however,
do not model the full posterior distribution but are a discrete approximation
of the Bayesian approach.

Blundell \etal~\cite{blundell2015bayesByBackprop} show that Bayesian methods
can be applied to neural networks to model probability distributions over the
trainable weights instead of point estimates. This property enables the network to predict uncertainty based on which the analyst can choose whether or not to trust the model's prediction.
Bayesian
neural networks have demonstrated impressive results in various tasks, \eg,
pixel-wise depth regression~\cite{kendall2017uncertaintiesInCV}, biomedical
image segmentation~\cite{kwon2018uncertainty}. We argue that this framework is
also particularly well suited for the field of forensics.

We demonstrate the usefulness of the Bayesian framework on the classical
forensic task of resampling detection. In resampling detection, the assumption
is that when an object is spliced into an image, it is likely resized or
rotated to fit into the target scene. Algorithmically, resizing or rotation is
typically implemented as a resampling operation.
In the past 15 years, many analytic forensic techniques were proposed to detect
resampling, e.g., via quasi-periodic inter-pixel
correlations~\cite{popescu2005exposing}, random matrix theory~\cite{vazquez2017random}, or natural
image statistics~\cite{goodall2016blind}.
Several learning-based methods achieve similar goals, either by directly
detecting resampling~\cite{flenner2018resampling} or the resampling
factor~\cite{liu2019cnn}, or indirectly via camera-based image forgery
localization~\cite{cozzolino2018camera} or the detection of image
splicing~\cite{yao2019improving}.
%
%
%
However, learning-based methods are sensitive to mismatches between training
and test data. For example, Liu and Kirchner report that their CNN for
rescaling factor estimation suffers from poor generalization to
unseen resampling algorithms like bicubic interpolation~\cite{liu2019cnn}.
They also report that including more diverse training examples lead to a
reduction in the overall accuracy. Moreover, it is certainly intractable to
cover all possible real-world scenarios in the training data.


\begin{figure}[t]
  \centering
    \includegraphics[width=1.\linewidth]{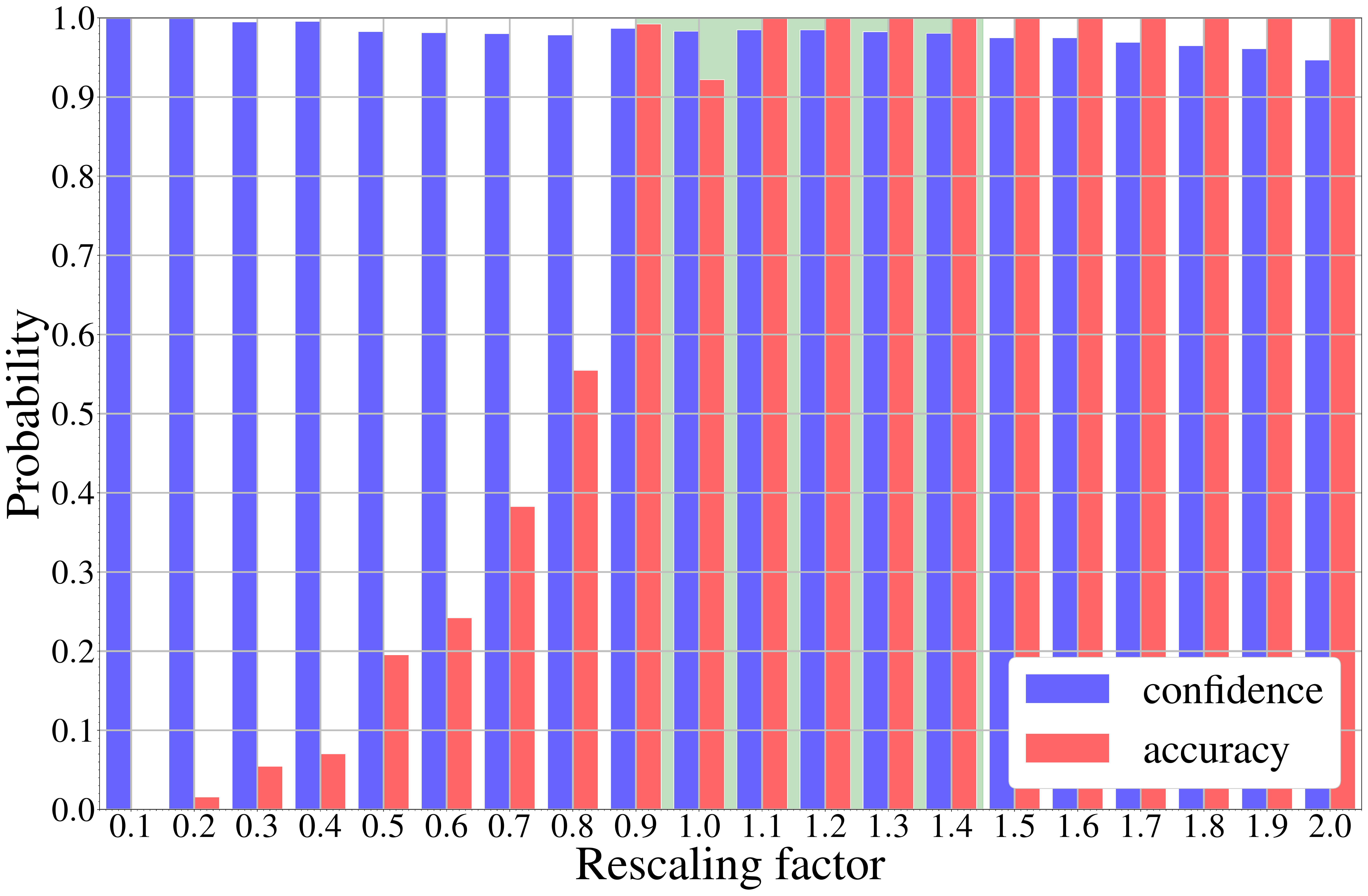}
    \caption{Baseline CNN: prediction confidence (blue) versus accuracy (red), averaged over $128$ samples for scaling factors from $0.1$ to $2.0$ with step size $0.1$ (see text for details).}
    \label{fig:results_standard_conf_vs_acc}
\end{figure}

\begin{figure*}[tb]
    \centering
    \centerline{\includegraphics[width=\textwidth]{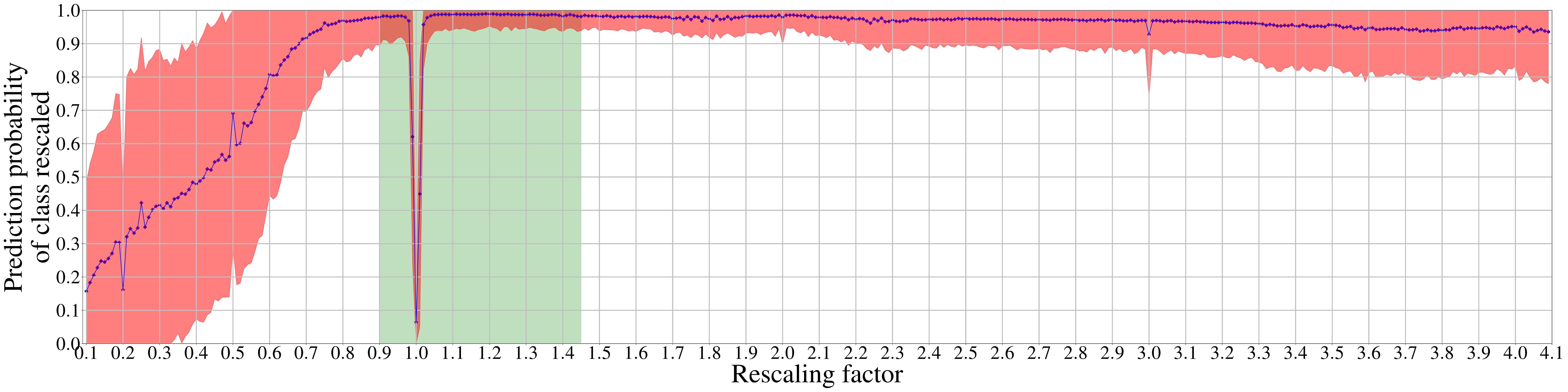}}
    
    \caption{Uncertainty of the Bayesian CNN. Blue: prediction probability for ``rescaled'', averaged over 128 input samples.
	Red: two standard deviations from the mean, calculated from $50$ Monte Carlo draws from the posterior probability.
	Green: scaling factors for network training.}
    \label{fig:results_bayesian_normal}
\end{figure*}

\section{Bayesian Nets and Variational Inference}
\label{sec:methods}

Standard neural networks can be seen as universal function approximators, able to represent an arbitrary function $G(\pmb{x})$ within an arbitrary small but fixed distance $\varepsilon$ by a learned function $g_{\pmb{\omega}}(\pmb{x})$,
\begin{equation}
    |G(\pmb{x}) - g_{\pmb{\omega}}(\pmb{x})| < \varepsilon\enspace.
    \label{uat_bound}
\end{equation}
Here, $\pmb{\omega} \in \rm I\!R^t$ denotes the $t$ trainable network parameters or weights.
Training is usually formulated as an optimization problem, and solved for the optimal weights $\pmb{\omega}$ via gradient descent.
These weights are scalar quantities and hence form a point estimate.  Standard
neural networks can learn powerful representations. However, they also suffer
from overly confident decisions on unseen data. Moreover, these models are by
design deterministic, and as such not able to express uncertainty in their
decisions.
%
%


The BNN does not learn point estimates, but instead the posterior
distribution
$P(\pmb{\omega}|\mathcal{D})$ over the weights $\pmb{\omega}$ given some
training data $\mathcal{D} = \{\left(\pmb{x}_i, \pmb{y}_i\right)\}, \; i= 1
\dots S$.
Here, $S$ is the number of training samples, and the samples are tuples of input
images $\pmb{x}_i$ and their corresponding class label $\pmb{y}_i$ that follows
a categorical distribution.
Exact Bayesian inference, i.e., the exact calculation of the posterior, is
intractable due to the large number of parameters in a neural network.
Blundell \etal~\cite{blundell2015bayesByBackprop} introduced the \textit{Bayes
by Backprop} algorithm, which approaches a variational approximation of the
posterior distribution.
It enables to learn a probability distribution over the trainable weights in a network.

Variational learning aims to find optimal parameters $\theta$ of the weight distribution $q(\pmb{\omega}| \theta)$,
which is achieved when $q(\pmb{\omega}| \theta)$ is similar to the true unknown distribution $P(\pmb{\omega}| \mathcal{D})$.
This corresponds to minimizing the Kullback-Leibler divergence
\begin{equation}
    \begin{aligned}
        \theta^{\ast} &= \argmin_\theta KL\left[q(\pmb{\omega}|\theta)||P(\pmb{\omega}|\mathcal{D})\right] \\
        &= \argmin_\theta \int q(\pmb{\omega}|\theta)\log{\frac{q(\pmb{\omega}|\theta)}{P(\pmb{\omega}| \mathcal{D})}}d\pmb{\omega} \\
        &= \argmin_\theta \int q(\pmb{\omega}|\theta)\log{\frac{q(\pmb{\omega}|\theta)}{P(\pmb{\omega})P(\mathcal{D}|\pmb{\omega})}}d\pmb{\omega} \\
        &= \argmin_\theta \int q(\pmb{\omega}|\theta)\log{\frac{q(\pmb{\omega}|\theta)}{P(\pmb{\omega})}} - q(\pmb{\omega}|\theta)\log{P(\mathcal{D}|\pmb{\omega})}d\pmb{\omega} \\
        &= \argmin_\theta \int q(\pmb{\omega}|\theta)\log{\frac{q(\pmb{\omega}|\theta)}{P(\pmb{\omega})}}d\pmb{\omega} - \int q(\pmb{\omega}|\theta)\log{P(\mathcal{D}|\pmb{\omega})}d\pmb{\omega} \\
        &= \argmin_\theta KL\left[q(\pmb{\omega}|\theta)||P(\pmb{\omega})\right] - \mathord{\rm I\!E}_{q(\pmb{\omega}|\theta)}\left[\log{P(\mathcal{D}|\pmb{\omega})}\right] \enspace.
    \end{aligned}
    \label{Bayes_opt_prob}
\end{equation}
In the last equation, the second term denotes the negative log-likelihood, which is usually optimized via maximum likelihood estimation (MLE).
The first term acts as regularizer in a maximum \emph{a posteriori} (MAP)
sense. Overall, Eqn.~\ref{Bayes_opt_prob} minimizes the negative
log-likelihood while enforcing a small Kullback-Leibler divergence between the
weight distribution and the prior distribution.
This cost function is known as the evidence lower bound (elbo) loss.
Direct optimization of this cost function is computationally expensive.
However, it can be approximated as a function of the training data
$\mathcal{D}$ and the variational parameters $\theta$ via gradient descent and
the dominated convergence theorem, which allows to interchange a derivative with
an expectation.
This allows to rewrite the optimization problem in Eqn.~\ref{Bayes_opt_prob} to
\begin{equation}
    f(\pmb{\omega}, \theta) = \log{q(\pmb{\omega}|\theta)} - \log{P(\pmb{\omega})} - \log{P(\mathcal{D}|\pmb{\omega})}\enspace.
\end{equation}
The exact cost then can be approximated as
\begin{equation}
    \mathcal{F}(\mathcal{D}, \theta) \approx \sum_{i=1}^{n}\log q(\pmb{\omega}^i|\theta) - \log P(\pmb{\omega}^i) - \log P(\mathcal{D}|\pmb{\omega}^i)\enspace,
\end{equation}
where $\pmb{\omega}^i$ denotes the $i$-th sample drawn from the variational posterior.
The Bayes by Backprop algorithm introduced by Blundell \etal~\cite{blundell2015bayesByBackprop} was formulated for feed-forward neural networks using fully connected layers.
Their approach can be generalized to convolutional neural networks by applying
flipout convolution~\cite{wen2018flipout}. This enables the estimation of the
predictive posterior via sampling from the variational
posterior, similar to~\cite{kwon2018uncertainty},
\begin{align}
    \mathord{\rm I\!E}_{q(\pmb{\omega}|\theta)}\left[P(\pmb{y}^*|\pmb{x}^*)\right] & = \int P(\pmb{y}^*|\pmb{x}^*, \pmb{\omega}) q(\pmb{\omega}|\theta) d\pmb{\omega} \\
	& \approx \frac{1}{n}\sum_{i=1}^n P_{\pmb{\omega}^i}(\pmb{y}^*|\pmb{x}^*)\enspace,
    \label{var_approx}
\end{align}
where $\pmb{x}^*$ denotes unseen data, $\pmb{y}^*$ the predicted class label and $P_{\pmb{\omega}^i}(\pmb{y^*}|\pmb{x}^*)$ a draw from the predictive posterior.
The approximation in Eqn.~\ref{var_approx} samples $n$ times from the trained network on unseen data. The variance of this estimator expresses the network's uncertainty in its prediction~\cite{kwon2018uncertainty},
\begin{equation}
    \mathrm{Var}(P(\pmb{y}|\pmb{x})) = \mathord{\rm I\!E}_{q(\pmb{\omega}|\theta)}[\pmb{y}\pmb{y}^T] - \mathord{\rm I\!E}_{q(\pmb{\omega}|\theta)}[\pmb{y}]\mathord{\rm I\!E}_{q(\pmb{\omega}|\theta)}[\pmb{y}]^T\enspace.
    \label{eq:variance_calc}
\end{equation}



\section{Experimental results}
\label{sec:experiments}

We conduct a series of experiments on rescaling detection to evaluate the
robustness of the Bayesian CNN (BNN) and its ability to express predictive
uncertainty. 

\textbf{Dataset Preparation.} 
We randomly select \numprint{1000} uncompressed
high-resolution images from the RAISE dataset~\cite{dang2015raise}. Each RGB
color image is converted to
grayscale using the ITU-R 601-2 luma transform. The \numprint{1000} images are
split into \numprint{800} images for training, \numprint{100} images for
validation, and \numprint{100} images for testing. Each image is further
processed as follows: one copy is left as original, and one copy is rescaled
with a rescaling factor $s = 0.9+k\cdot 0.05$, where $k$ is randomly chosen
between $0$ and $11$ excluding $2$, i.e., scaling factors are between
$0.9$ and $1.45$ excluding the identity $1.00$. From both copies we randomly draw $N=50$
non-overlapping patches of $256\times 256$ pixels. Hence,
training, validation, and testing sets consist of \numprint{80000},
\numprint{10000}, and \numprint{10000} patches from disjunct images. 


\textbf{Network Architectures.}
As a baseline, we use the popular constrained convolutional architecture by Bayar and
Stamm~\cite{bayar2017robustness}. We implement this architecture in Tensorflow, but omit the extremely
randomized trees classifier in order to perform a frictionless comparison of
end-to-end deep learning architectures.

The Bayesian network uses the baseline network as template, with the same
number of layers, same number and dimensions of the filter kernels per
convolution layer, and the same constrained convolution layer.
The Bayesian property is obtained via flipout convolution and fully-connected
layers~\cite{wen2018flipout} from the Tensorflow probability
framework~\cite{dillon2017tensorflow}.
As prior distribution we assume a zero-mean Gaussian distribution with unit-variance.
For inference, we use Eqn.~\ref{var_approx} with $n=50$ Monte Carlo draws, and
we calculate the predictive variance via Eqn.~\ref{eq:variance_calc}.
We assume a normally-distributed variational posterior, hence the Bayesian
network has twice as many training parameters as the baseline CNN. 

\textbf{Training Parameters}
%
We use the Adam optimizer with a learning rate of $l=10^{-3}$, $\beta_1 = 0.9$, $\beta_2 = 0.999$ and $\epsilon=10^{-7}$, and a batch size of $64$. 
The baseline is trained for \numprint{100000} iterations. The BNN for \numprint{150000} iterations due to twice the network parameters. The reported results use the model that best performs on the validation set, which is evaluated every \numprint{1000} iterations.

\subsection{Detection Accuracy}
The baseline model and the Bayesian CNN are trained for the detection of
rescaling using the same datasets and hyper-parameters. On the test set, the
baseline achieves $96.32\%$ accuracy. This is comparable
to~\cite{bayar2017robustness} given that the additional extremely
randomised trees for performance boosting are omitted. The Bayesian CNN achieves
$97.40\%$ accuracy, which is comparable, even slightly better.

\begin{figure}[t!]
    \centering
    \begin{subfigure}[t]{0.5\linewidth}
        \centering
        \includegraphics[width=\textwidth]{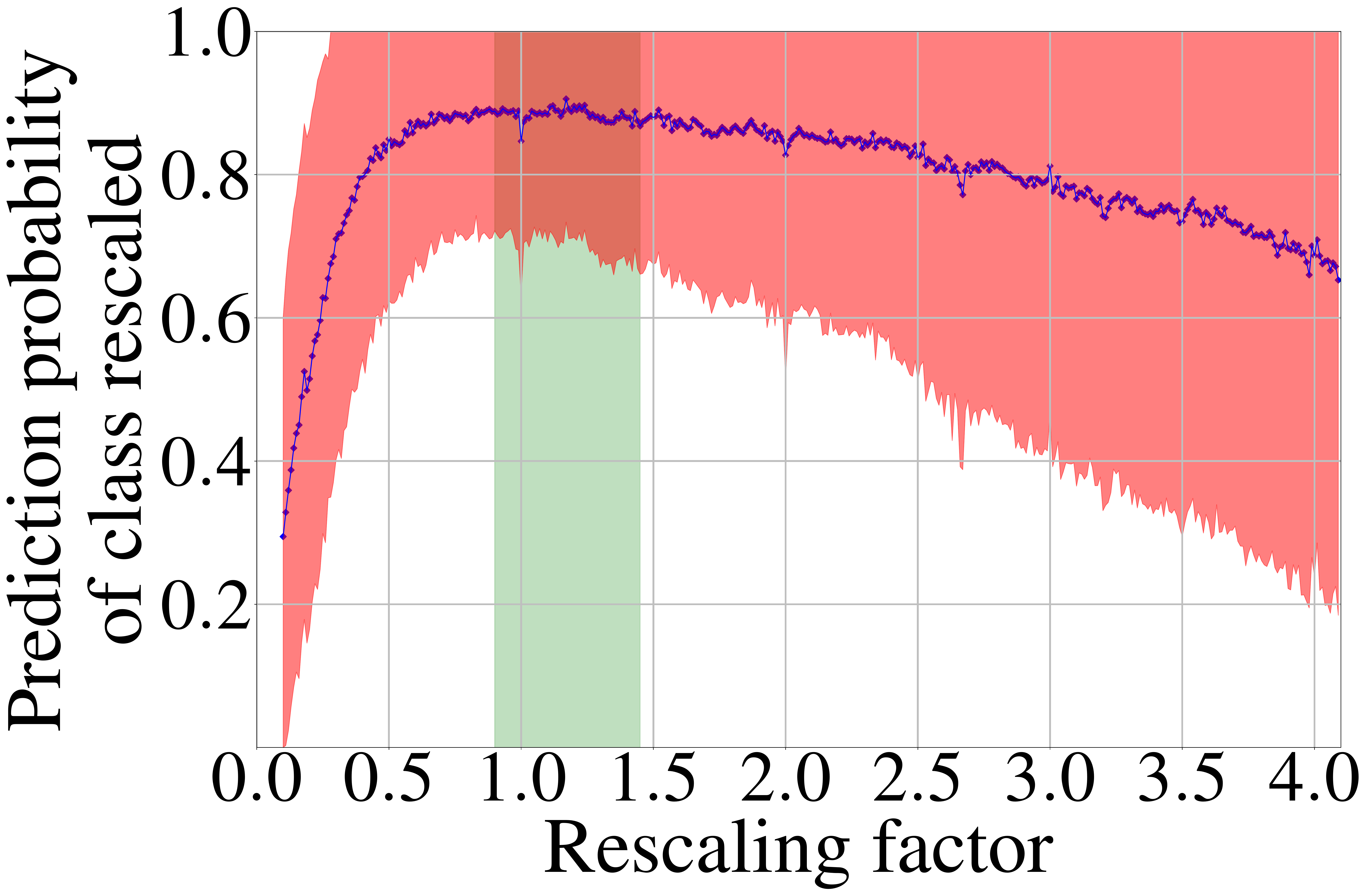}
        \caption{JPEG-compression ($q=85$)}
    \end{subfigure}%
    \hfill
    \begin{subfigure}[t]{0.5\linewidth}
        \centering
        \includegraphics[width=\textwidth]{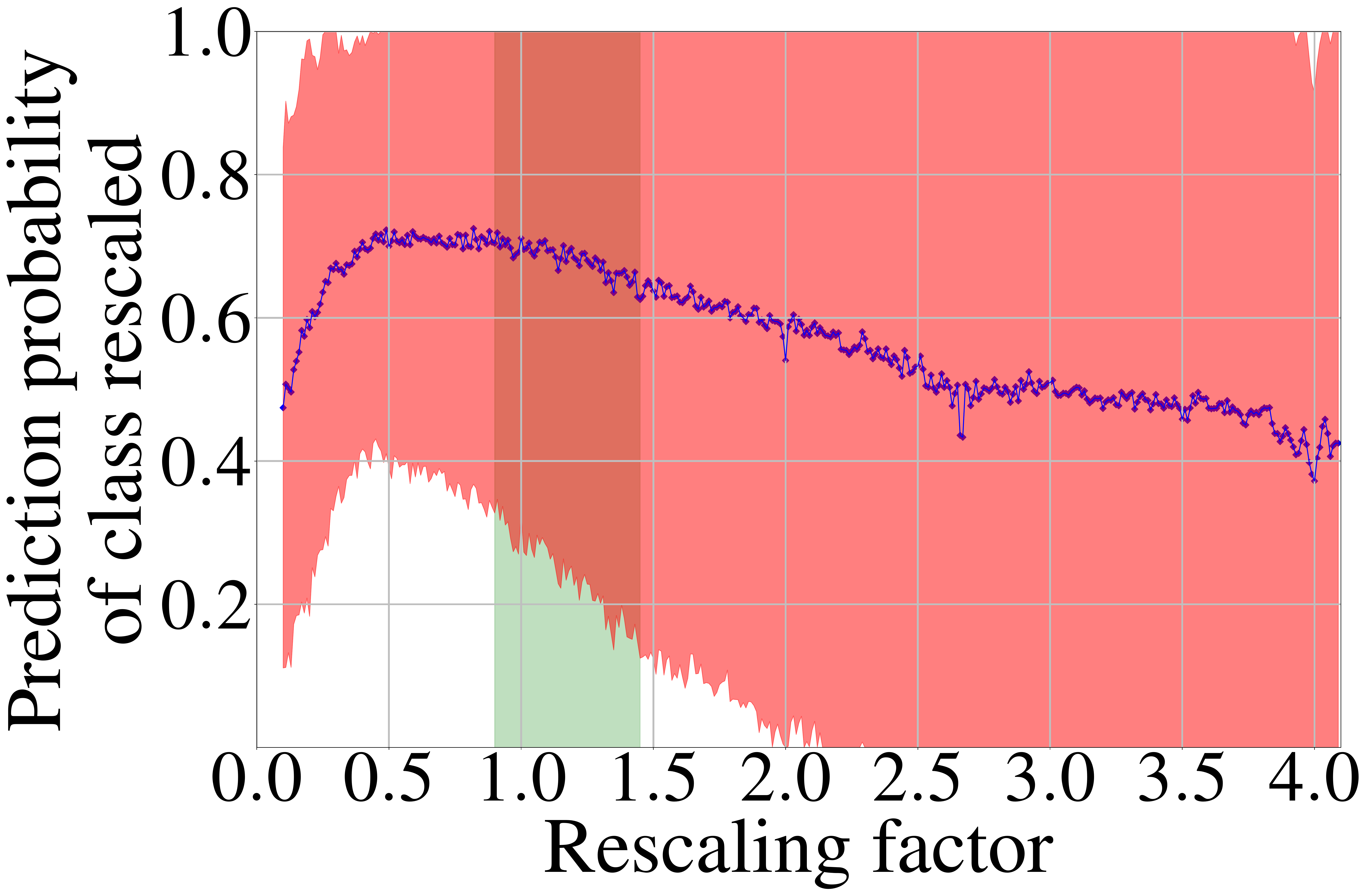}
        \caption{JPEG-compression ($q=50$)}
    \end{subfigure}
    \caption{Uncertainty by the Bayesian CNN on out-of-distribution test samples with JPEG postcompression.}
    \label{fig:result_bayes_out_of_dist_jpeg}
\end{figure}

\subsection{Standard CNN and Out-of-Distribution Samples} 

We first show that a standard CNN does not provide helpful information to
detect out-of-distribution samples. To this end, recall that the baseline CNN
is trained on resampling factors in the range of $s \in \{0.9 \dots 1.45\}$. 
For testing,
%
%
we create a second test set with resampling factors $s' = \{0.1 \dots 2.0\}$ in
steps of $0.1$. With exception of the resampling range, the test set
preparation follows the exact same protocol as the previous dataset. In
particular, the test images are unseen during network training.
We select $M=128$ patches per rescaling factor, and calculate
average accuracy and confidence per rescaling factor. The confidence is
calculated as~\cite{hendrycks17softmaxStatisticsBaseline}
\begin{equation}
c_s = \frac{1}{M}\sum_{m=1}^{M}\max_{y_k} P(y_k \mid \pmb{x}_m)\enspace,
\end{equation}
i.e., averaging the highest activation per decision per rescaling factor.

The results are shown in Fig.~\ref{fig:results_standard_conf_vs_acc}.
The rescaling factors $0.1$ to $2.0$ are on the $x$-axis.  The range of
rescaling factors for network training is shown in green.  Red bars indicate
the accuracy of the baseline method per rescaling factor $s$. Blue bars
indicate the associated confidence $c_s$.  It can be observed that the blue
confidence bars are always very high, at or beyond $0.9$.
Moreover, the confidence is independent of the actual network accuracy, which
significantly varies across resampling factors. This makes it impossible to
predict the network performance from the distribution of class activations.


\subsection{Uncertainty in Out-of-Distribution Resampling Factors}

The previous experiment is repeated with the BNN. To this end, we expand
the range of rescaling factors even further to $[0.1; 4.1]$ with 
step size $0.01$, and leave all other experimental parameters identical.

The results are shown in Fig.~\ref{fig:results_bayesian_normal}. 
Green indicates the range of training data.
The blue line shows the mean prediction probability for ``rescaled'', averaged
over $M=128$ randomly selected patches.
Red indicates two standard deviations of the uncertainty.
The network excellently performs inside the train region, distinguishing the
original and rescaled patches with over $99\%$ confidence, while exhibiting
very low uncertainty in its decision.
The network generalizes very well for scaling factors $s \geq 1.5$.
With increasing distance to the training region, the uncertainty grows
accordingly with a slight reduction in accuracy, which is the desired behavior.
For downscaling, the network is able to detect scaling factors $s \geq 0.7$,
i.e., in direct proximity to the train region.
For scaling factors $s < 0.7$, the network prediction performance drops
considerably, and the uncertainty again grows accordingly.
Hence, the network uncertainty indicates its inability to operate on this input.  We
consider this an important cue for a forensic analyst to not trust the
network output.
%
%

Both networks, the baseline in Fig.~\ref{fig:results_standard_conf_vs_acc} and
the Bayesian CNN in Fig.~\ref{fig:results_bayesian_normal}, make errors on
out-of-distribution examples. However, where the standard CNN makes extremely
overconfident errors, the Bayesian CNN tends to be uncertain, which can inform
the analyst whether she can trust the network prediction or not.

\begin{figure}[t!]
    \centering
    \begin{subfigure}[t]{0.5\linewidth}
        \centering
        \includegraphics[width=\textwidth]{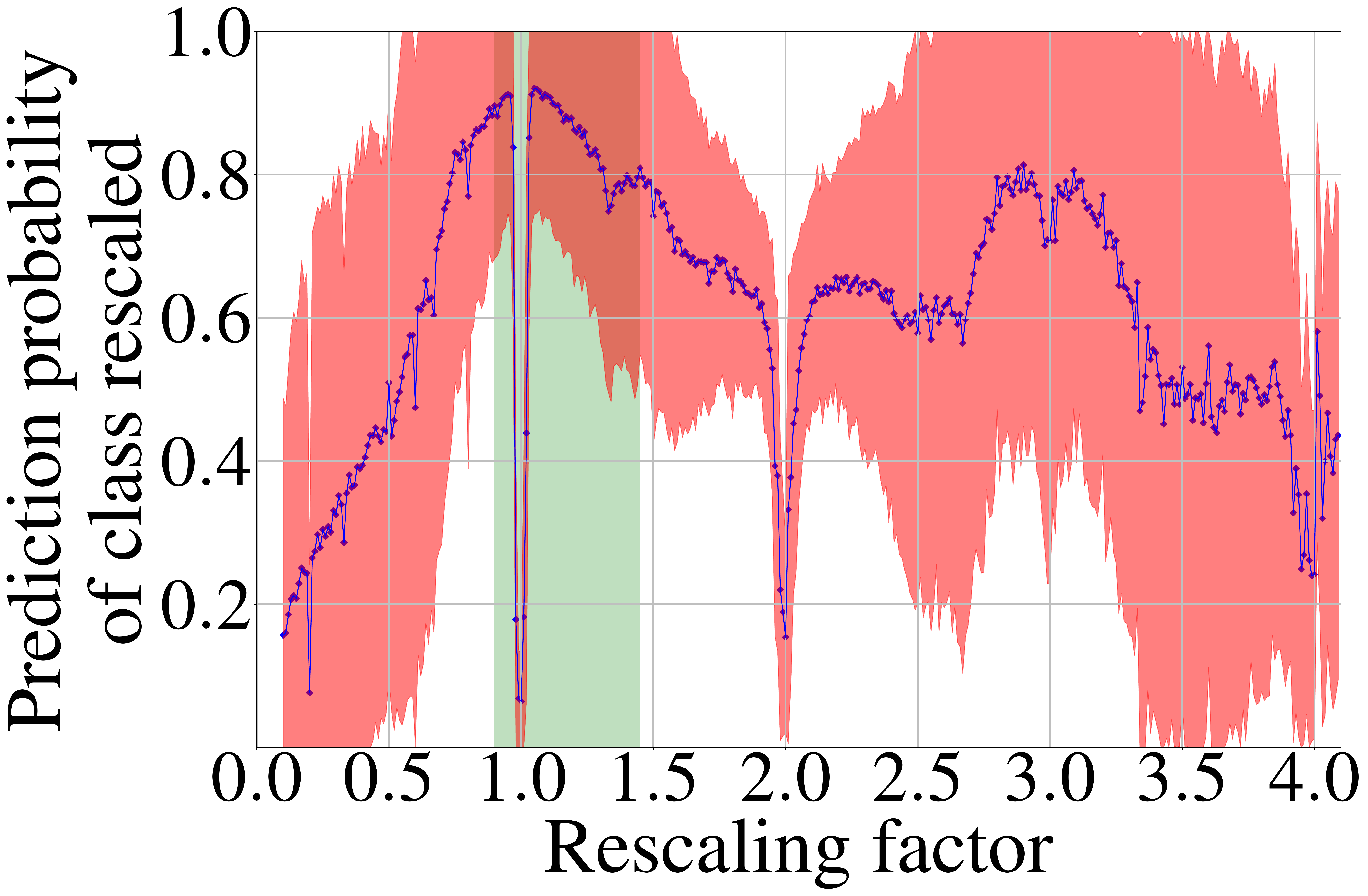}
        \caption{nearest-neighbor interpolation}
    \end{subfigure}%
    \hfill
    \begin{subfigure}[t]{0.5\linewidth}
        \centering
        \includegraphics[width=\textwidth]{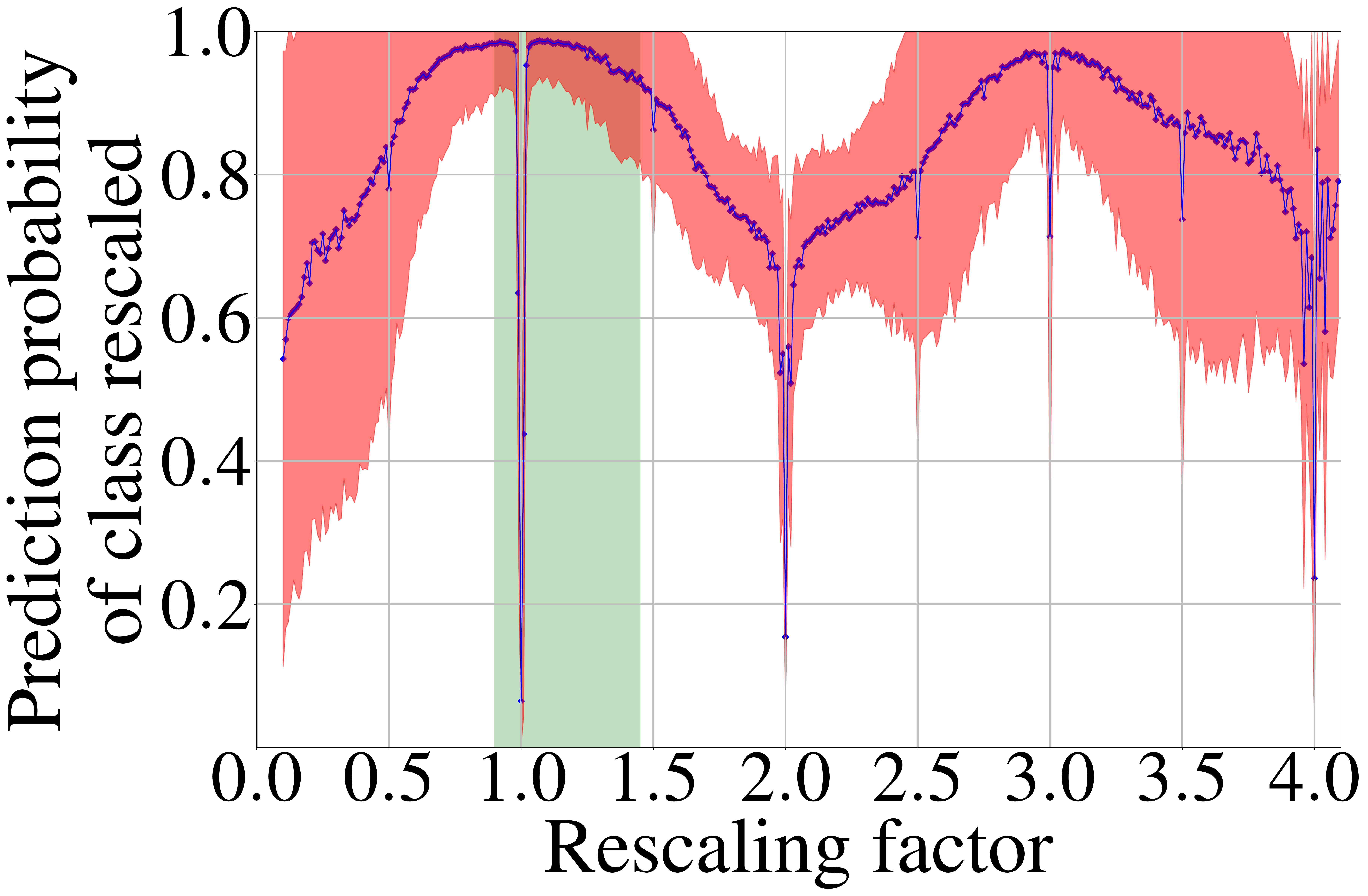}
        \caption{areal interpolation}
    \end{subfigure}
    \caption{Uncertainty by the Bayesian CNN on out-of-distribution test samples of nearest-neighbor and areal interpolation.}
    \label{fig:result_bayes_out_of_dist_inter}
\end{figure}

\subsection{Uncertainty in Out-of-Distribution JPEG Compression}
There are numerous scenarios that might not be well covered in the classifier
training set. We select two cases, namely unseen resampling methods and JPEG
compression after resampling.

Data augmentation with JPEG compression is routinely performed in many forensic
algorithms. Nevertheless, we believe that JPEG compression is an interesting
experiment, for it is widely accepted in the forensic community as a prototypic
case for out-of-distribution samples: since the BNN training data is
uncompressed, we can observe how the BNN uncertainty serves as a metric for the
network prediction reliability.
In this experiment, we recompress all testing data with quality factors $q=85$
and $q=50$. The remaining experimental protocol is identical to the previous
sections.

The results of this experiment are shown in
Fig.~\ref{fig:result_bayes_out_of_dist_jpeg}. As expected, the prediction
probability drops with increasing compression and with increasing deviation
from the training resampling parameters. It is encouraging to observe that this
is accurately indicated by a simultaneous increase in the uncertainty.

\subsection{Uncertainty in Out-of-Distribution Resampling Operations}

%

A more subtle case of out-of-distribution samples are variations in the
resampling operations. Such deviations in the data distribution are much more
difficult to catch, and might even be non-obvious to a technical expert.  In
this experiment, we use nearest neighbor interpolation and an areal
interpolation on the testing data.
The remaining experimental protocol is identical to the
previous sections.

The results of this experiment are shown in
Fig.~\ref{fig:result_bayes_out_of_dist_inter}.
Analogous to the JPEG experiments, prediction probabilities are also
significantly lower, but the uncertainty increase accordingly, which again
makes it possible to detect the mismatch between training and testing data.

\section{Conclusion}
\label{sec:conclusion}

We propose a novel Bayesian deep learning approach to express predictive
uncertainty in image forensics. We show on the example of resampling detection
that the Bayesian CNN avoids confident errors. Moreover, the Bayesian framework
intrinsically provides an uncertainty estimate that indicates a model mismatch
to a forensic analyst, which is otherwise difficult to recognize.

This is preliminary work. We believe that BNNs can close an important gap in
image forensics, but there are still many aspects to be investigated. In future
work, we will extend this approach to other forensic tasks, explore selection
strategies for prior distributions, and explore a decomposition of the
predictive uncertainty into model and data
uncertainty~\cite{kwon2018uncertainty}.

%


\bibliographystyle{IEEEbib}
\bibliography{strings,refs}

\end{document}